\documentclass[11pt]{article}

\usepackage[preprint]{acl}

\usepackage{times}
\usepackage{latexsym}

\usepackage[T1]{fontenc}

\usepackage[utf8]{inputenc}

\usepackage{microtype}

\usepackage{inconsolata}

\usepackage[utf8]{inputenc}

\usepackage{graphicx}
\usepackage{subcaption}
\usepackage{caption}
\usepackage{pdflscape}

\usepackage{booktabs}  
\usepackage{adjustbox}

\title{Investigating Query-Insensitive Behavior in \\Spatio-Temporal Video Grounding}

\author{Eryk Kołodziejczyk$^1$, Alberto Presta$^1$, Karol Szurkowski$^1$, Michał Byra$^{1,2}$ \\
\\
 \textsuperscript{1}Samsung AI Center, Warsaw, Poland, \\
 \textsuperscript{2}Institute of Fundamental Technological Research,
 \\Polish Academy of Sciences, Warsaw, Poland
\\
 \small{
   \textbf{Correspondence:} \href{mailto:e.kolodziejc@samsung.com}{e.kolodziejc@samsung.com}}
}

\begin{document}

\maketitle
\begin{abstract}
Spatio-temporal video grounding (STVG) aims to localize objects or events described by natural language queries in both space and time. Existing STVG models are typically trained and evaluated under the assumption that each query is relevant to the input video. In this work, we challenge this assumption by studying the behavior of state-of-the-art STVG models under irrelevant queries and missing textual input. Our experiments show that current models can still produce plausible spatio-temporal predictions even when the query is unrelated to the video or removed entirely. We further analyze HCSTVG-v2 and VidSTG to identify dataset regularities that may encourage such query-insensitive behavior. Our study highlights an underexplored limitation of STVG models and motivates negative-aware evaluation protocols and architectures that explicitly assess query relevance.
\end{abstract}

\section{Introduction}

The rapid growth of video content on the internet has underscored the need for efficient event localization methods. Spatio-temporal video grounding (\textbf{STVG}) has emerged as a key research area at the intersection of natural language processing and computer vision, combining text query understanding with video understanding to detect the desired video fragment. 

This study examines the sensitivity of STVG models to queries unrelated to any event or object in the video. Trained exclusively on positive query-video pairs, these models often generate false positive predictions for irrelevant queries. Notably, such predictions frequently align with plausible regions or temporal segments, suggesting reliance on dataset-specific patterns rather than query-video relevance. Our analysis of prominent STVG datasets reveals inherent biases in their distributions, which may hinder robust generalization.

Although negative awareness has recently been studied in related tasks, such as visual-language reasoning and video moment retrieval \cite{asadi2026mirage,flanagan2025momentuntruth}, it remains underexplored in STVG, especially with respect to spatial predictions. This work addresses the gap in negative awareness for STVG by: (1) analyzing model behavior under irrelevant queries, exposing limitations in fully supervised approaches, and (2) examining the relation between predictions and dataset distributions to identify biases impeding robust model development.

\section{Related Work}
\paragraph{STVG.} Spatio-temporal video grounding bridges natural language understanding with fine-grained video localization. Early work in this area formulated STVG as a pipeline of smaller subtasks \cite{yamaguchi2017spatio}. The introduction of annotated datasets like VidSTG \cite{zhang2020vidstg} and HCSTVG-v2 \cite{tang2021hcstvg}, accelerated model development. Fully supervised methods, such as TubeDETR~\cite{yang2022tubedetr}, STCAT~\cite{jin2022embracingconsistency}, CG-STVG~\cite{gu2024context} and TA-STVG~\cite{gu2025tastvg}, directly learn from annotated video-query pairs. These methods have substantially improved STVG performance, but they are typically trained under the assumption that each query is relevant to the video.

\paragraph{Query sensitivity.} Recent works examined model behavior under invalid or irrelevant inputs. Mirage~\cite{asadi2026mirage} demonstrates that vision-language models can achieve high scores on benchmarks even with empty image inputs. Similarly, Moment of Untruth~\cite{flanagan2025momentuntruth}, identifies a critical flaw in temporal video grounding (TVG): the assumption that queried moments exist in videos, leading to false positives for irrelevant queries. To mitigate this, authors follow negative rejection works \cite{pudil1992recreject, herbei2006rejectclass, wiener2011agnosticclass, chen2023benchmarkllmrag, franc2023rejectood} and add a classification head to assess query-video relevance, requiring training data with both positive and negative queries. Another publication, Temporal Sentence Grounding with Relevance Feedback~\cite{dong2024relfeedback}, proposes a new task that extends temporal video grounding to handle cases where query-related segments may not exist in a video, and proposes a Relation-aware Temporal Sentence Grounding (RaTSG) network that performs foreground-background detection at multiple granularities to localize segments or confirm their absence. Learning to Refuse~\cite{lee2025learningrefuse}, proposes reinforcement fine-tuning to filter out irrelevant queries in TVG. 

In contrast, our work investigates query sensitivity in STVG, where irrelevant queries impact both temporal and spatial grounding. We conduct an in-depth analysis of STVG models behavior under irrelevant queries, characterize the resulting false positive predictions, and examine the composition of prominent STVG datasets to identify factors contributing to this phenomenon.

\section{Methods}

\subsection{Spatio-temporal video grounding}

Spatio-temporal video grounding aims to localize the object or event described by a natural language query in both space and time. Given a video and a text query, the model predicts a spatio-temporal tube $STT = \{b_t\}_{t=t_s}^{t_e}$, where $b_t$ denotes the bounding box at time $t$, and $t_s$ and $t_e$ denote the start and end frames of the predicted temporal segment.

In this work, we analyze two fully supervised STVG models, TubeDETR and TA-STVG~\cite{yang2022tubedetr,gu2025tastvg}. TubeDETR, an early transformer-based model, employs a video-text encoder for spatial multimodal interactions and a space-time decoder for spatio-temporal localization. TA-STVG, representing the current state-of-the-art, enhances TubeDETR with text-guided temporal sampling and attribute-aware spatial activation, improving both spatial and temporal accuracy. These models correspond to key advancements in fully supervised STVG approaches.

\subsection{Datasets and training}
We use two established STVG datasets for our experiments. VidSTG dataset \cite{zhang2020vidstg} is  based on the VidOR and YFCC100M datasets \cite{shang2019vidor, thomee2016yfcc100m}, and contains 6,924 videos divided into 99,943 video-text samples. There are two types of queries, 44,808 declarative and 55,135 interrogative sentences. Data samples are divided into 80,684 training, 8,956 validation and 10,303 test subsets. HCSTVG-v2 dataset \cite{tang2021hcstvg} includes 16,544 video-text pairs, divided into 10,131 training, 2,000 validation and 4,413 test subsets. Since test set is not annotated, we follow the previous work and report results on the validation set \cite{yang2022tubedetr,gu2025tastvg}. 

We evaluate STVG models using implementations from official repositories, adapting the training setups to match our GPU resources. We report performance using standard STVG metrics, including mean temporal Intersection over Union (m\_tIoU) and mean video Intersection over Union (m\_vIoU), which measure spatio-temporal overlap between the predicted and ground-truth tubes. Details are provided in the Appendix.

\section{Results}

\subsection{Characterization of datasets}

In this section, we characterize the datasets and identify potential sources of bias in STVG models.

\begin{figure*}[!t]
\centering
    \captionsetup[subfigure]{justification=centering}
    \begin{subfigure}[b]{0.23\textwidth}
        \includegraphics[width=\columnwidth]{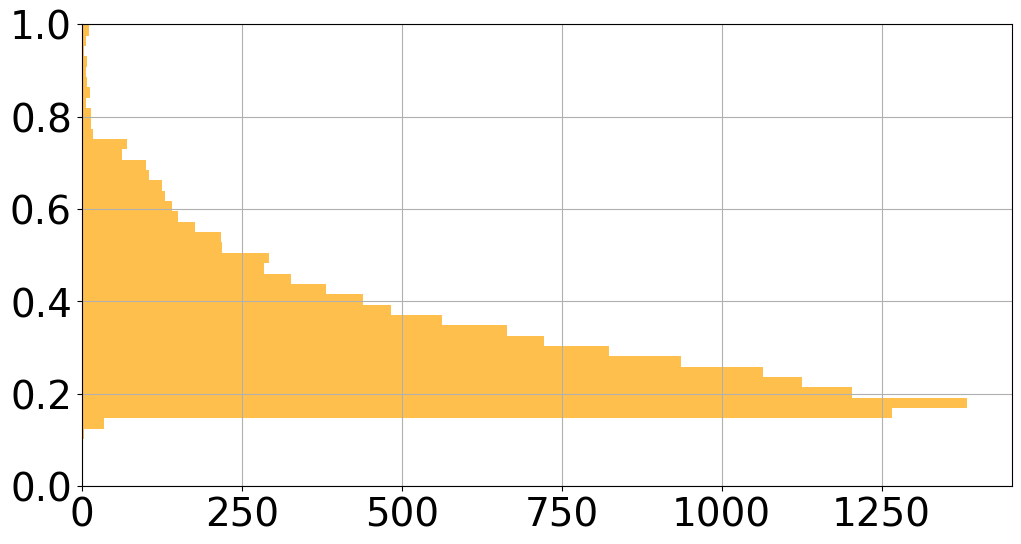}
        \caption{HCSTVG-v2\\moments duration}
        \label{fig:sub1}
    \end{subfigure}
    \hfill
    \begin{subfigure}[b]{0.23\textwidth}
        \includegraphics[width=\columnwidth]{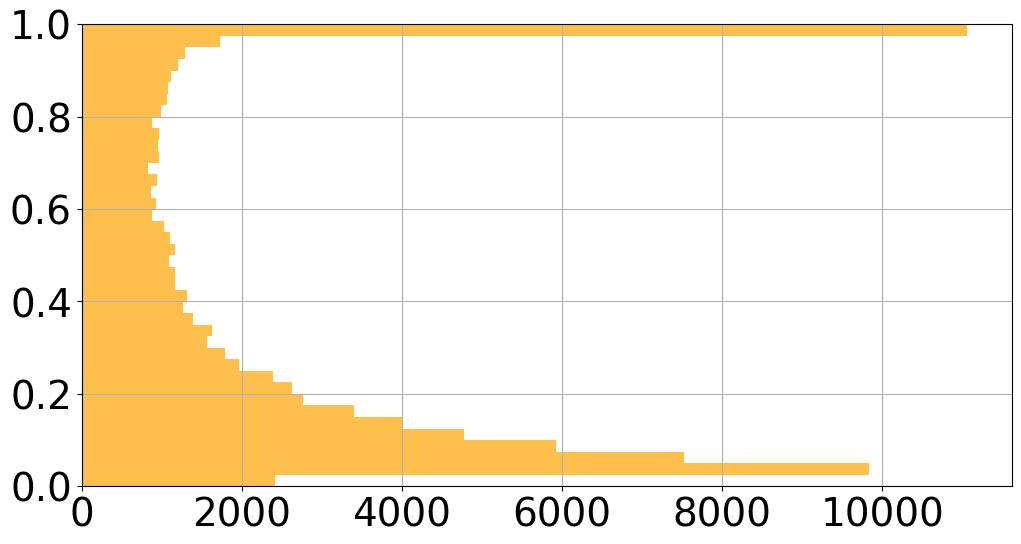}
        \caption{VidSTG\\moments duration}
        \label{fig:sub2}
    \end{subfigure}
    \hfill
    \begin{subfigure}[b]{0.23\textwidth}
        \includegraphics[width=\columnwidth]{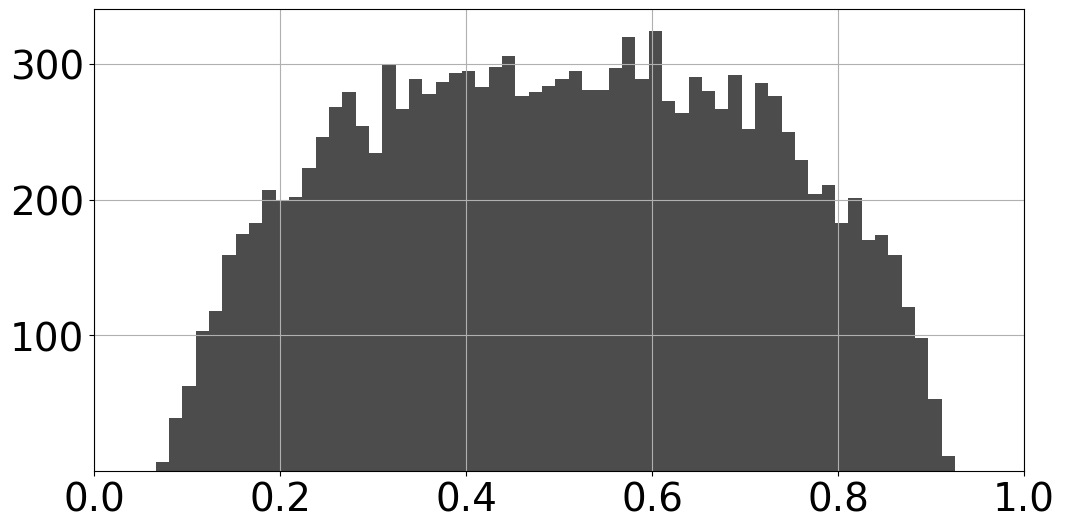}
        \caption{HCSTVG-v2\\moments centers}
        \label{fig:sub3}
    \end{subfigure}
    \hfill
    \begin{subfigure}[b]{0.23\textwidth}
        \includegraphics[width=\columnwidth]{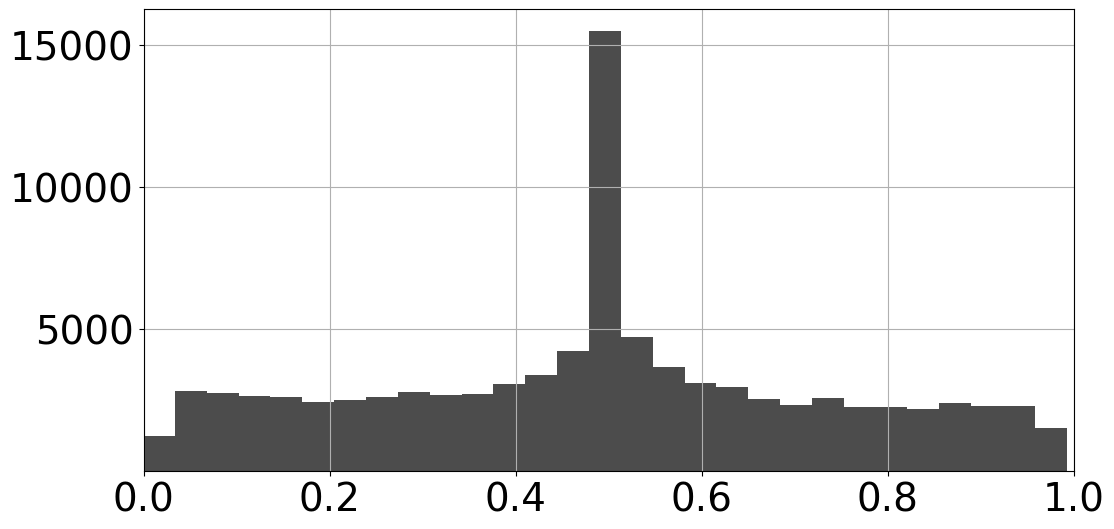}
        \caption{VidSTG\\moments centers}
        \label{fig:sub4}
    \end{subfigure}
\caption{Temporal analysis of HCSTVG-v2 and VidSTG: moment duration and its time center. Values are decimal units normalized to the video duration.}
\label{fig:temporal_analysis}
\end{figure*}

\paragraph{Temporal analysis.} The datasets differ substantially in temporal structure. VidSTG contains nearly ten times more samples than HCSTVG-v2 and includes clips ranging from 3 to 180 seconds, while all HCSTVG-v2 samples last 20 seconds. As depicted in Fig.~\ref{fig:temporal_analysis}, annotated moments in HCSTVG-v2 usually span 20--40\% of the video. In VidSTG, they most often cover either a short fragment of 2.5--10\% or the entire video. Analysis of the normalized moment midpoint reveals a broad distribution in HCSTVG-v2, while VidSTG is strongly concentrated around the video center, partly due to many full-video annotations.

\begin{figure}
\centering
\begin{subfigure}{0.48\columnwidth}
    \includegraphics[width=\columnwidth]{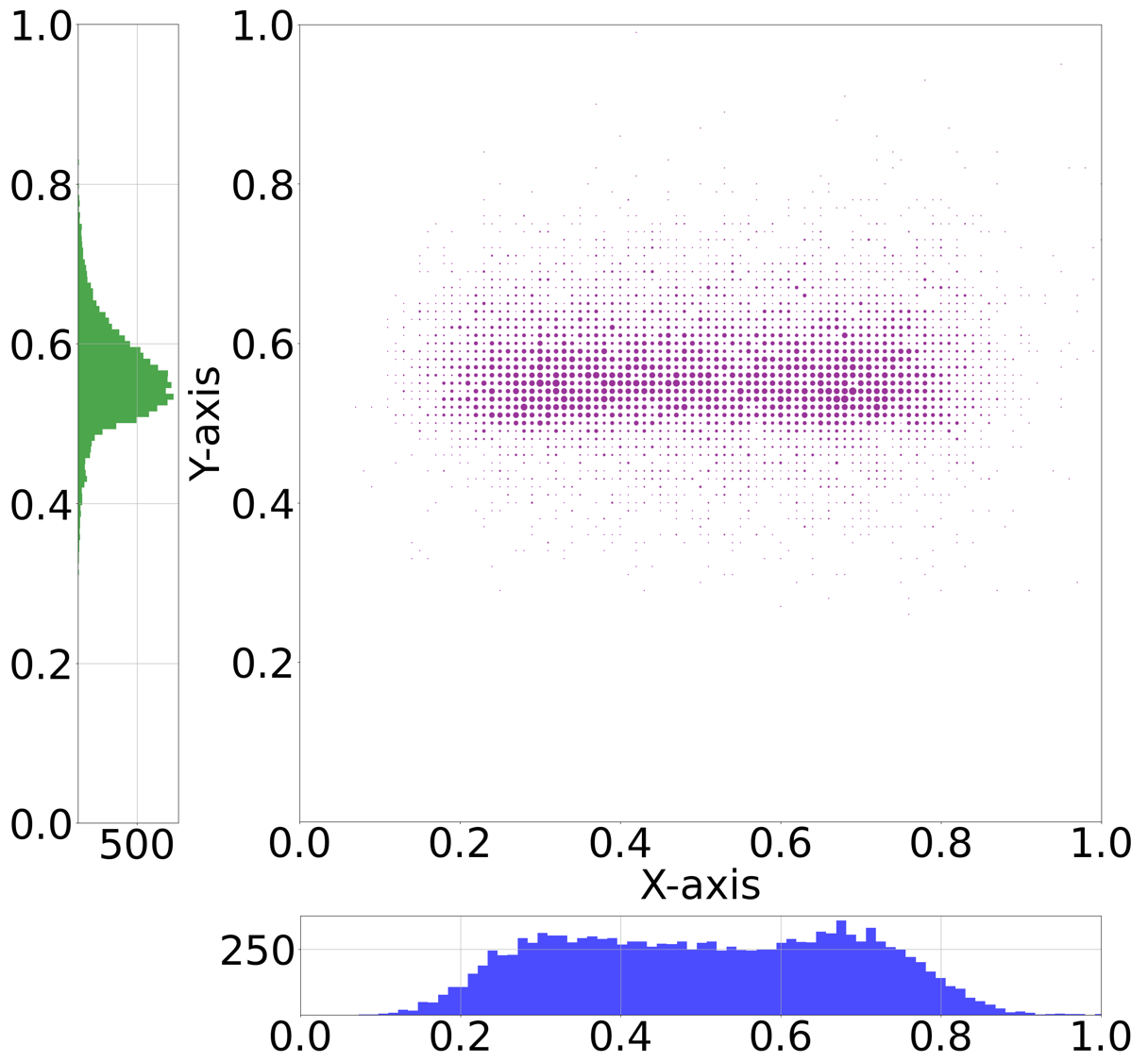}
    \caption{HCSTVG-v2}
    \label{fig:spat_first}
\end{subfigure}
\hfill
\begin{subfigure}{0.48\columnwidth}
    \includegraphics[width=\columnwidth]{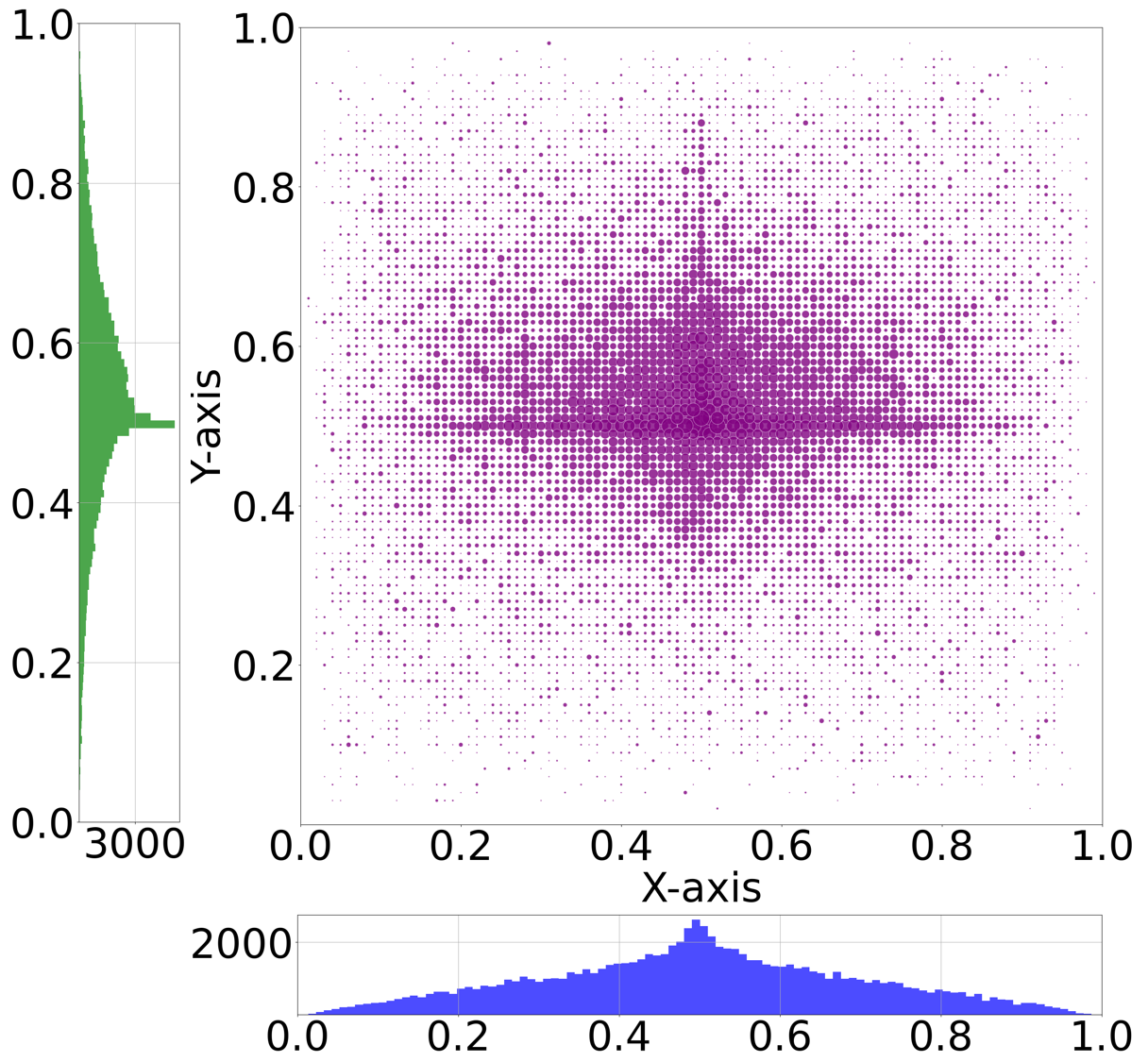}
    \caption{VidSTG}
    \label{fig:spat_second}
\end{subfigure}

\caption{Comparison of spatial annotations between HCSTVG-v2 and VidSTG datasets. Datapoints show average bounding box centers on X and Y axes, normalized to video frame size. Scatter plot indicates frequency of centers at the same spot.}

\label{fig:spatial_analysis}
\end{figure}

\paragraph{Spatial analysis.} We next analyze spatial annotation patterns in the two datasets. Since STVG requires frame-level bounding boxes within each annotated moment, we summarize each sample using its average bounding box and mean box area. Figure~\ref{fig:spatial_analysis} reveals a pronounced vertical bias in both datasets, with bounding box centers clustered near the frame center. Horizontally, VidSTG also shows a central bias, while HCSTVG-v2 has a more uniform distribution. This pattern may reflect common recording practices, where the object of interest is often kept near the center of the scene.

\begin{figure}[!t]
\centering
\begin{subfigure}[b]{\columnwidth}
    \includegraphics[width=\columnwidth]{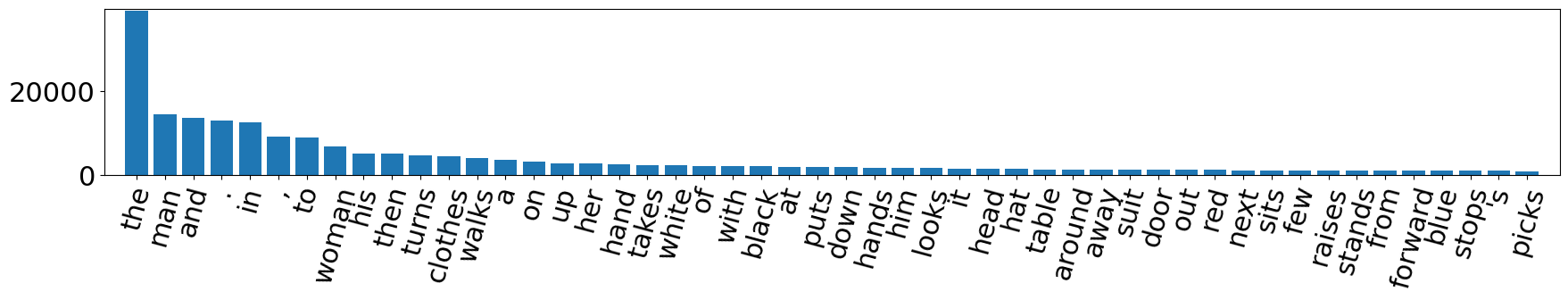}
    \caption{HCSTVG-v2}
    \label{fig:subw1}
\end{subfigure}

\vspace{4pt}

\begin{subfigure}[b]{\columnwidth}
    \includegraphics[width=\columnwidth]{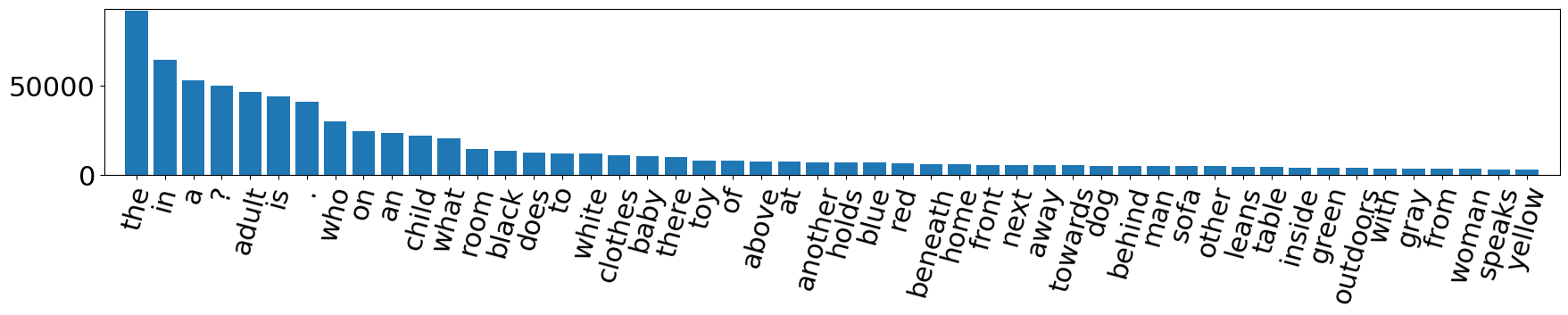}
    \caption{VidSTG}
    \label{fig:subw2}
\end{subfigure}
\caption{Comparison of words distribution in datasets.}
\label{fig:queries-analysis}
\end{figure}

\paragraph{Query analysis.} To analyze the textual content of video datasets, we conducted a word frequency analysis of the queries using NLTK (v. 3.9.1). 
Figure \ref{fig:queries-analysis} illustrates the distribution of words across the STVG datasets. In HCSTVG-v2, the most frequent terms include words \textit{'the'} and \textit{'man'}, with \textit{'woman'} appearing significantly less frequently. Conversely, VidSTG exhibits a higher prevalence of \textit{'adult'} and introduces terms like \textit{'children'} and \textit{'babies'}, which are less common in HCSTVG-v2. 
Verb analysis reveals that HCSTVG-v2 emphasizes action-oriented terms such as \textit{'turns'}, \textit{'walks'}, \textit{'takes'}, \textit{'puts'} and \textit{'looks'}, whereas VidSTG frequently uses \textit{'is'}, focusing on descriptive rather than action-based queries.

\paragraph{Summary.} 
The analysis identifies the following dataset biases. HCSTVG-v2, being human-centric with fixed 20-second samples, may lead models to rely on temporal structures and human-centered cues. VidSTG, while larger and more diverse, also exhibits strong temporal and spatial regularities, such as short or full-video annotations and centrally located bounding boxes. These patterns risk encouraging models to exploit dataset-specific priors rather than grounding queries in video content.


\subsection{Sensitivity to negative queries}
Here, we analyze the behavior of TubeDETR and TA-STVG~\cite{yang2022tubedetr,gu2025tastvg} under negative queries. These queries are irrelevant to the input video, but the model is still required to produce a spatio-temporal prediction. We use the authors' released checkpoints for the best-performing configurations to assess model behavior.

\begin{table*}[!t]
\centering
\resizebox{\textwidth}{!}{%
\begin{tabular}{c|cccc|cccc|cccc}
\toprule
 & \multicolumn{4}{c|}{\textbf{HCSTVG-v2}}
 & \multicolumn{4}{c|}{\textbf{VidSTG declarative}}
 & \multicolumn{4}{c}{\textbf{VidSTG interrogative}}  \\
 & \multicolumn{2}{c|}{\textbf{TubeDETR}}
 & \multicolumn{2}{c|}{\textbf{TA-STVG}}
 & \multicolumn{2}{c|}{\textbf{TubeDETR}}
 & \multicolumn{2}{c|}{\textbf{TA-STVG}}
 & \multicolumn{2}{c|}{\textbf{TubeDETR}}
 & \multicolumn{2}{c}{\textbf{TA-STVG}} \\
\textbf{Query type}
 & \textbf{ m\_vIoU} & \textbf{m\_tIoU}&\textbf{ m\_vIoU} & \textbf{m\_tIoU}
 & \textbf{ m\_vIoU} & \textbf{m\_tIoU}&\textbf{ m\_vIoU} & \textbf{m\_tIoU}
 & \textbf{ m\_vIoU} & \textbf{m\_tIoU}&\textbf{ m\_vIoU} & \textbf{m\_tIoU}\\ 
\midrule
Positive
 & 0.343 & 0.525 & 0.335 & 0.548
 & 0.283 & 0.468 & 0.342 & 0.517
 & 0.244 & 0.461 & 0.291 & 0.501 \\
\midrule
Negative in-domain
 & 0.170 & 0.416 & 0.206 & 0.476
 & 0.083 & 0.298 & 0.097 & 0.314
 & 0.092 & 0.318 & 0.101 & 0.323 \\

 & -50\% & -21\% & -39\% & -13\%
 & -71\% & -36\% & -72\% & -39\%
 & -62\% & -31\% & -65\% & -36\% \\
\midrule
Negative out-of-domain
 & 0.161 & 0.398 & 0.186 & 0.457
 & 0.078 & 0.273 & 0.106 & 0.274
 & 0.074 & 0.276 & 0.085 & 0.279 \\

 & -53\% & -24\% & -44\% & -17\%
 & -73\% & -42\% & -69\% & -47\%
 & -70\% & -40\% & -71\% & -44\% \\
\bottomrule
\end{tabular}%
}
\caption{\label{tab:neg_inference}
Inference results: we evaluate TubeDETR and TA-STVG on VidSTG and HCSTVG-v2 datasets, using m\_vIoU and m\_tIoU. Three query types are compared: positive, negative in-domain, and negative out-of-domain. For negative queries, we report relative IoU changes compared to positive ones. A smaller decrease in the IoU value indicates a behavior that is less sensitive to changes in the query.}
\end{table*}


We follow the methodology of Moment of Untruth~\cite{flanagan2025momentuntruth} and evaluate two types of negative queries. \textbf{In-domain negatives} are from the same dataset but assigned to an unrelated video.
\textbf{Out-of-domain negatives} describe categories absent in the dataset, such as a physics laboratory. Query generation method, with examples, is described in the Appendix. For reference, we also assess the models using the original positive queries. 

In Table~\ref{tab:neg_inference}, we compare model predictions for different query types. For all query types, including negative queries, we compute the scores against the ground-truth annotations of the original positive query. Therefore, high scores are expected for positive queries, whereas high scores for negative queries indicate false positive grounding. The models perform best with original queries. However, some negative queries yield comparable scores, showing that STVG models can still localize the original annotated region even when the input query is irrelevant to the video. This behavior follows from the standard STVG formulation, which assumes query-video relevance and forces the model to output temporal segments and spatial boxes for input queries. The effect is more pronounced in HCSTVG-v2, where IoU drops are minimal, while VidSTG shows smaller effects. We observed that for negative cases the outputted bounding boxes usually corresponded to main entities in the video. In addition, the relative gap between positive and negative queries is smaller for m\_tIoU than for m\_vIoU, indicating a larger issue in temporal localization than in spatial grounding.

\subsection{Sensitivity to textual input}

The prevalence of false positives for irrelevant queries suggests that STVG models may prioritize salient regions over query-specific localization. To examine this possibility, we assess TA-STVG's performance when textual input is removed or limited, evaluating its reliance on text for grounding.

\subsubsection{Training with empty queries}

To ensure a fair comparison, we train the model both with the original queries and with empty queries under the same setup. Our reproduced results differ from those reported by the authors, which we discuss in the Appendix. Standard deviations are computed over three training runs.  

Table~\ref{tab:neg_train} compares models trained on original and modified datasets (empty queries) for HCSTVG-v2 and VidSTG. Validation results show minimal m\_tIoU differences, while m\_vIoU gaps are larger, indicating stronger textual input influence on spatial grounding. HCSTVG-v2 exhibits a more pronounced lack of negative awareness.

\begin{table}
  \centering
  \resizebox{\columnwidth}{!}{%
    \begin{tabular}{l|cc|cc}

    \midrule
    & \multicolumn{2}{c|}{\textbf{HCSTVG-v2}}
     & \multicolumn{2}{c}{\textbf{VidSTG}} \\
    \midrule
    \textbf{Query}&\textbf{ m\_vIoU} & \textbf{m\_tIoU}&\textbf{ m\_vIoU} & \textbf{m\_tIoU}\\
    \midrule
    Original & ${0.357}_{\pm{0.01}}$ & ${0.535}_{\pm{0.00}}$ & ${0.267}_{\pm{0.02}}$ & ${0.468}_{\pm{0.01}}$ \\
    \midrule
    Empty & ${0.262}_{\pm{0.01}}$ & ${0.485}_{\pm{0.01}}$  & ${0.151}_{\pm{0.02}}$ & ${0.399}_{\pm{0.00}}$ \\
     & -27\% & -9\% & -43\% & -15\%\\
  \end{tabular}
  }
  \caption{Training validation results of TA-STVG model on HCSTVG-v2 and VidSTG datasets. Comparison of trainings with original and empty query. For empty query, we report IoU changes relative to original one.}
  \label{tab:neg_train}
\end{table}

\subsubsection{Training without text modules}

We remove TA-STVG's text-processing components (e.g. text encoder), leaving only the video-processing modules. The model is trained without any textual guidance, while using the same grounding losses as the original architecture. We conduct this experiment on HCSTVG-v2 because the previous analysis indicated stronger query-insensitive behavior on this dataset. Moreover, HCSTVG-v2 provides a single query per video, making it a suitable setting for testing whether the model can exploit dataset regularities without textual input. Results in Table~\ref{tab:no_text_encoder_train} show m\_vIoU dropped from 0.357 to 0.238 and m\_tIoU from 0.535 to 0.479. This indicates the model learns dataset-driven biases without queries, with temporal localization more affected than spatial grounding. Our findings suggest that STVG models may exploit dataset regularities for the grounding.

\begin{table}
  \centering
  \resizebox{\columnwidth}{!}{%
  \begin{tabular}{l|cc}
    \hline
    \textbf{Model version} & \textbf{m\_vIoU} &\textbf{ m\_tIoU} \\
    \midrule
    TA-STVG, baseline & ${0.357}_{\pm{0.01}}$ & ${0.535}_{\pm{0.00}}$ \\
    \midrule
    TA-STVG, text-processing modules removed &  ${0.238}_{\pm{0.01}}$ & ${0.479}_{\pm{0.00}}$ \\
  \end{tabular}
  }
  \caption{Results for TA-STVG model trained on HCSTVG-v2 without the text-processing modules.}
  \label{tab:no_text_encoder_train}
\end{table}

\subsubsection{Query-blind center-prior baseline}

As our analysis in Sec. 4.1 shows strong spatial and temporal regularities in both datasets, natural question to ask is how do they impact the predictions. To verify it, we added a query-blind center-prior baseline. The baseline predicts a fixed temporal segment and a fixed bounding box estimated from training-set annotation statistics, and we evaluate it on the same splits as in the previous experiments.

For HCSTVG-v2, the predicted temporal segment is centered at the middle of the video and spans 20\% of its duration. The predicted box is centered at normalized coordinates $(0.5, 0.55)$
and has width 0.30 and height 0.75, estimated from the average training-set box statistics. For VidSTG, the predicted temporal segment is also at the middle of the video and spans 10\% of its duration. The predicted box is centered at $(0.5, 0.5)$ and has width 0.20 and height 0.40. All quantities are normalized by frame size and video duration.

As shown in Table~\ref{tab:distribution_experiment}, the center-prior baseline obtains nonzero scores, confirming that dataset-level annotation priors contribute to residual IoU. However, its performance is substantially lower than that of TA-STVG trained with empty queries. On HCSTVG-v2, the center-prior baseline obtains m\_vIoU of 0.063 and m\_tIoU of 0.248, compared with 0.262 and 0.485, respectively, for the empty-query model. On VidSTG, the center-prior baseline obtains 0.014 m\_vIoU and 0.089 m\_tIoU, while the empty-query model obtains 0.151 and 0.399. Unlike the center-prior baseline, the empty-query model still processes the video. Its substantially higher performance therefore indicates that the residual localization cannot be explained solely by global annotation statistics and also reflects video-conditioned visual cues. Nevertheless, the nontrivial center-prior scores, particularly for temporal localization on HCSTVG-v2, confirm that dataset regularities contribute to the observed behavior.

These results show that a trivial constant center prediction explains only partially the residual performance. The evaluated STVG models still obtain substantially higher IoU under irrelevant queries than this query-blind prior. The dataset priors contribute to the observed behavior, but do not fully account for it.

\begin{table}
  \centering
   \resizebox{\columnwidth}{!}{
      \begin{tabular}{l|cc|cc}
        \hline
         & \multicolumn{2}{c|}{\textbf{HCSTVG-v2}} & \multicolumn{2}{c}{\textbf{VidSTG}} \\ 
         & \textbf{m\_vIoU} &\textbf{ m\_tIoU} & \textbf{m\_vIoU} &\textbf{ m\_tIoU} \\
        \midrule
        Center-prior baseline & 0.063 & 0.248 & 0.014 & 0.089 \\
      \end{tabular}
    }
  \caption{Results from testing evaluation with constant spatio-temporal annotations taken from data distribution charts.}
  \label{tab:distribution_experiment}
\end{table}

\section{Conclusion}

We examined negative awareness in STVG models by analyzing their behavior under irrelevant queries and missing text inputs. The results showed that STVG models could produce plausible predictions even for invalid queries, revealing a limitation of STVG formulations. Dataset biases, such as centrally concentrated spatio-temporal annotations, can reinforce this query-insensitive behavior.

These findings highlighted the need for STVG evaluation protocols that include negative queries and explicitly measure a model’s ability to reject irrelevant text, rather than only localization accuracy on positive pairs. Future work should develop more diverse datasets, incorporate query-validity estimation, and design architectures or training objectives, such as contrastive learning \cite{chen2020contrastivelearning}, which increase dependence on textual input.

\section*{Limitations}
This work has several limitations. First, our experiments were conducted on two STVG models, which we consider representative of leading approaches on established benchmarks. Second, we analyzed two datasets that, to the best of our knowledge, are the primary benchmarks used for spatio-temporal video grounding. Third, our negative-query evaluation considered two types of queries: in-domain negatives, sampled from the same dataset but assigned to unrelated videos, and out-of-domain negatives, describing scenarios absent from the dataset. What is more, such reassigned queries cannot be guaranteed to be impossible in every target video, especially for generic human-centric descriptions. Other strategies of negative queries generation, such as random reassignment, semantic filtering, adversarial negatives and manually curated negatives, could be explored in future work. Finally, although we trained TA-STVG, we needed to adjust the training procedure because the model did not converge with the default configuration under our computational setup.
 
\label{sec:bibtex}
\bibliography{custom}

\appendix

\section{Comparison of spatial annotations in datasets - heatmaps}
\label{app:heatmaps}
\begin{figure}[!t]
\begin{subfigure}{0.45\textwidth}
    \includegraphics[width=\textwidth]{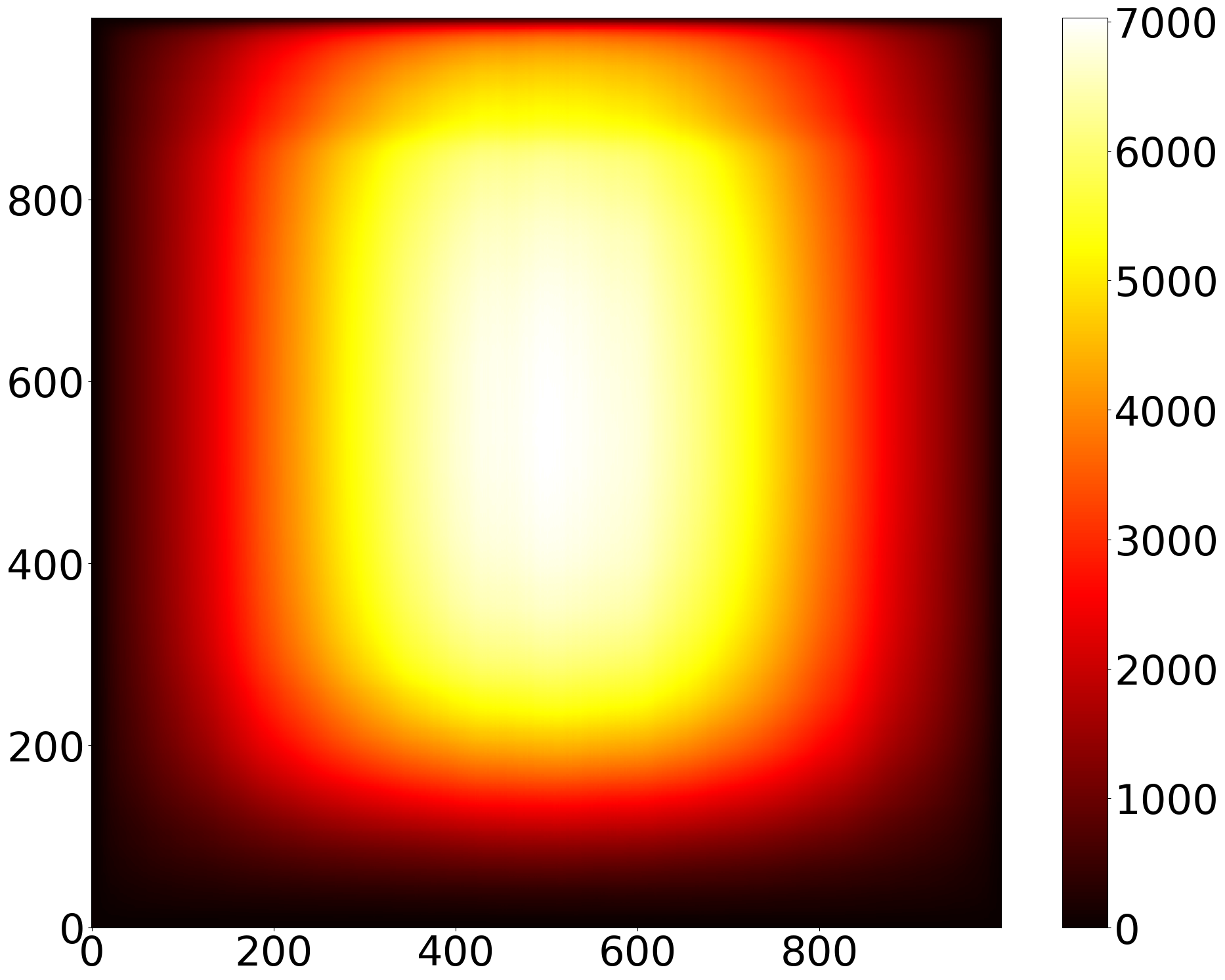}
    \caption{HCSTVG-v2}
    \label{fig:first}
\end{subfigure}
\begin{subfigure}{0.45\textwidth}
    \includegraphics[width=\textwidth]{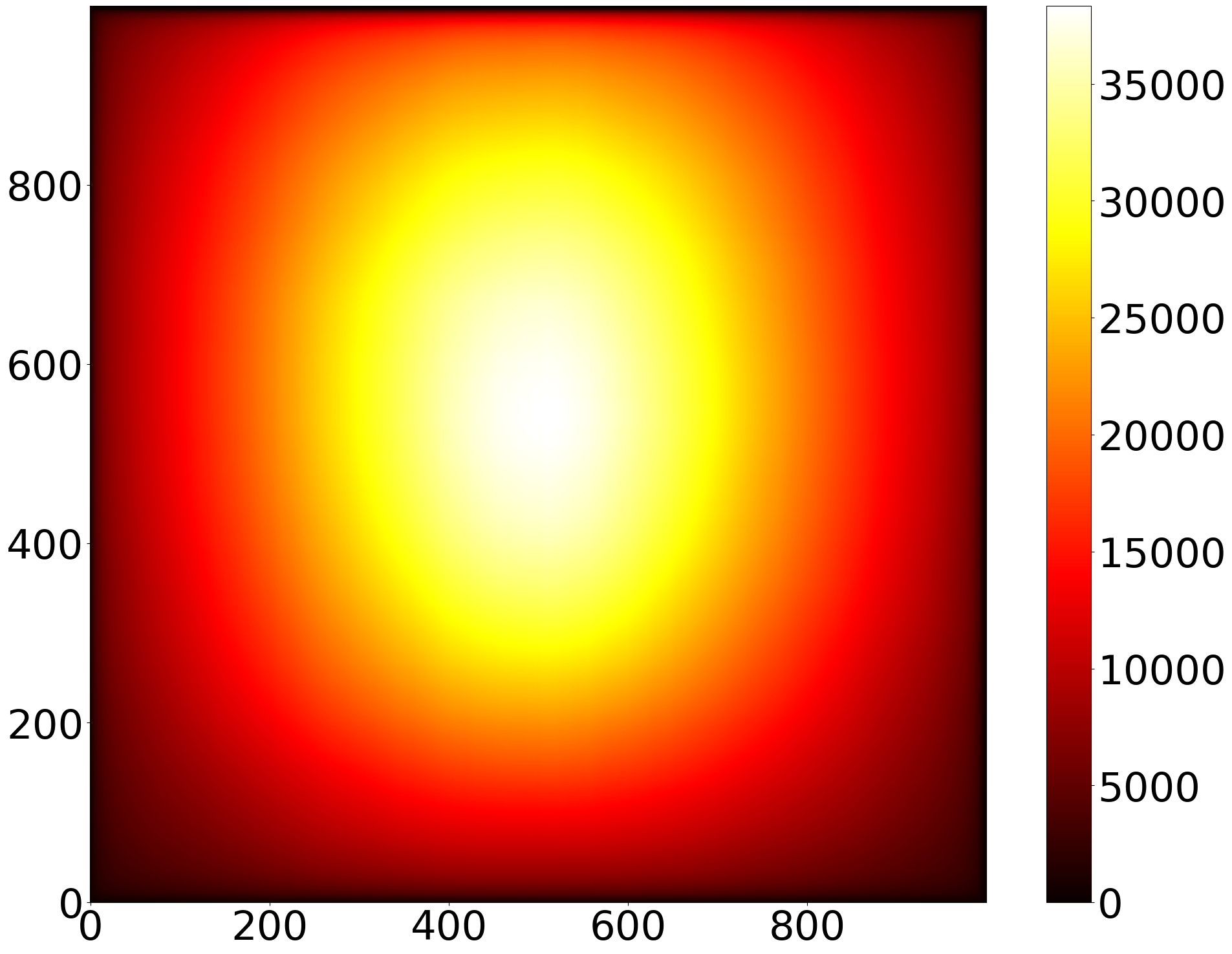}
    \caption{VidSTG}
    \label{fig:second}
\end{subfigure}
        
\caption{Comparison of averaged spatial heatmaps between HCSTVG-v2 and VidSTG dataset. Heatmaps illustrate mean bounding boxes of each video in the whole dataset.}
\label{fig:heatmaps}
\end{figure}

To better understand differences in spatial annotations between STVG datasets, we provide an additional visualization of bounding boxes. For each video, we compute the mean bounding box over time and normalize it by the video frame size, allowing annotations to be compared across the entire dataset. We then aggregate these normalized boxes into heatmaps, shown in Figure~\ref{fig:heatmaps}. The HCSTVG-v2 heatmap forms a vertically elongated region around the frame center, whereas VidSTG exhibits a more elliptical distribution. Moreover, VidSTG boxes cover a larger portion of the frame, while HCSTVG-v2 annotations leave the frame boundaries largely empty.

\section{Negative query generation}
\label{app:negative_queries}

We followed the negative-query construction strategy of Flanagan et al. \cite{flanagan2025momentuntruth}, adapting it from video moment retrieval to spatio-temporal video grounding. We considered two types of negative queries: in-domain and out-of-domain negatives.

\paragraph{In-domain negatives.}
In-domain negatives were obtained by reassigning existing queries from the same dataset to different videos. This preserves the linguistic style, object categories, and action vocabulary of the dataset, while removing the original query-video correspondence. Following Flanagan et al. \cite{flanagan2025momentuntruth}, we did not generate new in-domain sentences, as generated queries could accidentally describe content that is actually present in the target video. Instead, each negative query was sampled from existing annotations and paired with an unrelated video using the same methodology as in the original paper. This setting creates hard negatives, since the query remains plausible within the dataset domain but should not be grounded in the paired video.

\paragraph{Out-of-domain negatives.}
For out-of-domain negatives, we used the negative queries released by Flanagan et al. \cite{flanagan2025momentuntruth}. These queries were originally generated with LLMs from scenarios unlikely to appear in the evaluated video grounding datasets, using broad external domains and several subtopics within each domain. Examples of such domains include physics laboratories and mathematics classes. The following queries serve as samples: \textit{'the researcher downloads the satellite data to study the atmospheric composition'}, \textit{'she records the temperature and humidity readings in her lab notebook'} or \textit{'with a protractor, he measured the exterior angle of the polygon'}. We paired these queries with videos from the STVG datasets. In contrast to in-domain negatives, these queries evaluate whether a model can reject text that is clearly outside the visual and semantic scope of the video dataset.

\paragraph{Other negative queries generation methods}
As mentioned in the Limitations section, other methods of negative queries generation could be tested in the future studies. Different strategies are possible, including random reassignment, semantic filtering, LLM-generated out-of-domain negatives and manually curated negatives. Random reassignment is simple and scalable, but may occasionally create accidental positives. Semantic filtering can reduce this risk, but depends on the quality of the similarity model. LLM-generated out-of-domain negatives are less likely to accidentally match the video, but may be easier to reject because they are far from the dataset distribution. Manually curated negatives provide stronger guarantees, but are expensive to obtain.

\paragraph{Evaluation with negative queries.}
For both negative-query types, no valid ground-truth tube exists in the paired video. However, standard STVG models are formulated to always output a spatio-temporal tube for any input query. We therefore used negative queries only at inference time to analyze whether the models still produced plausible temporal segments and spatial boxes despite the absence of query-video correspondence.

\section{TA-STVG training}
\label{app:training}

We followed the official TA-STVG implementation and training protocol where possible \cite{gu2025tastvg}. TA-STVG uses ResNet-101 and RoBERTa-base initialized from MDETR as the 2D visual and text backbones, and VidSwin-tiny as the 3D motion backbone, with the 3D backbone frozen during training. The model samples 64 video frames, uses a maximum text length of 30 tokens, and applies random resized cropping with the output short side set to 420. The hidden dimension is 256 and the number of attention heads is 8. The original implementation uses Adam with learning rates of $2 \times 10^{-5}$ for the pre-trained backbones and $3 \times 10^{-4}$ for the remaining modules. Since the original training was performed on 32 A100 GPUs, while our experiments used 4 H100 GPUs, we adjusted the learning rates to account for the smaller effective batch size. Specifically, we used learning rate of $1 \times 10^{-5}$ for the entire model.

\section{Qualitative results}
\label{app:qualitative}

\begin{figure*}[ht]
\centering
    \begin{subfigure}[b]{1.0\textwidth}
        \centering
        \includegraphics[width=1\textwidth]{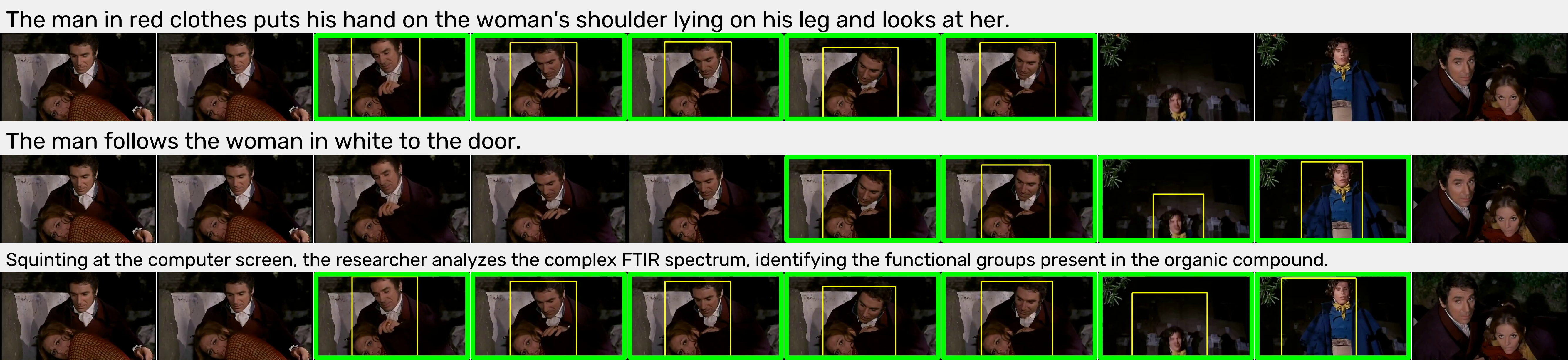}
        \caption{Video ID: 65\_G5Yr20A5z\_Q}
    \end{subfigure}
    \begin{subfigure}[b]{1.0\textwidth}
        \centering
        \includegraphics[width=1\textwidth]{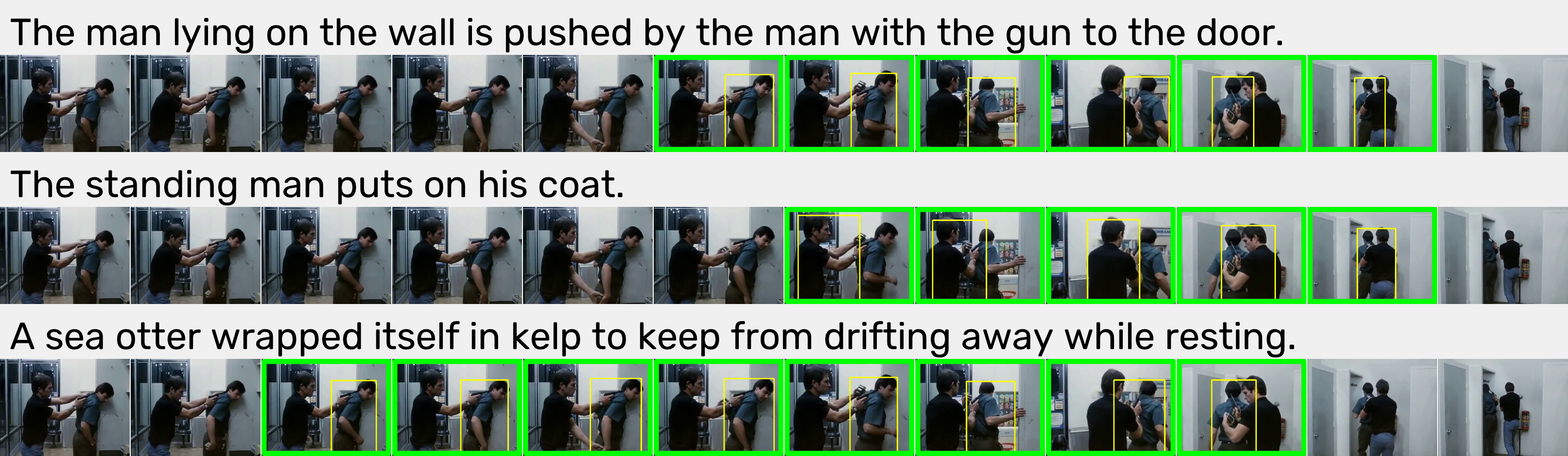}
        \caption{Video ID: 66\_oD\_wxyTHJ2I}
    \end{subfigure}
    \begin{subfigure}[b]{1.0\textwidth}
        \centering
        \includegraphics[width=1\textwidth]{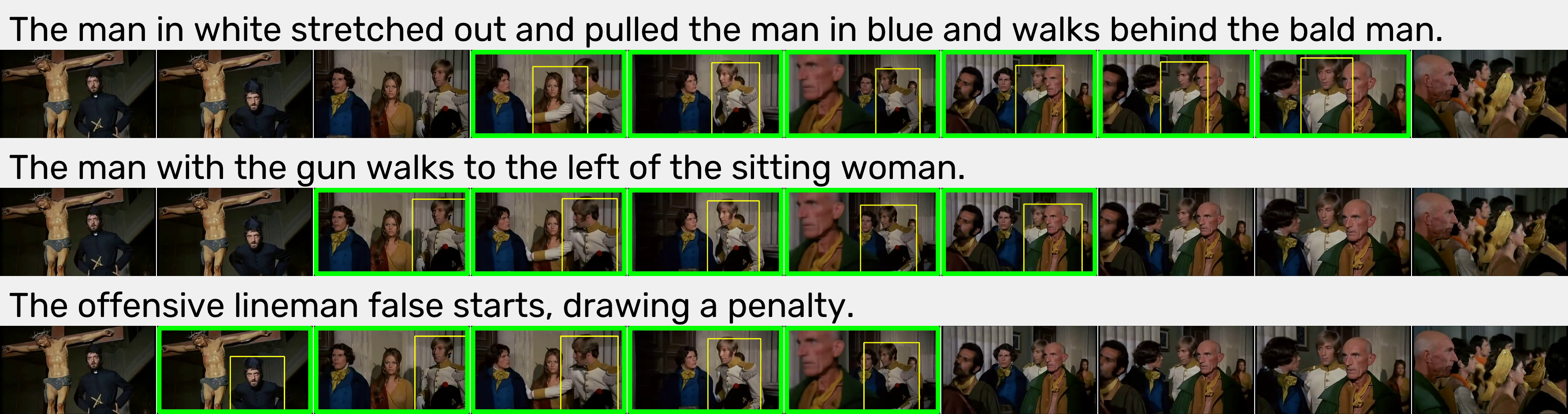}
        \caption{Video ID: 112\_G5Yr20A5z\_Q}
    \end{subfigure}
    \begin{subfigure}[b]{1.0\textwidth}
        \centering
        \includegraphics[width=1\textwidth]{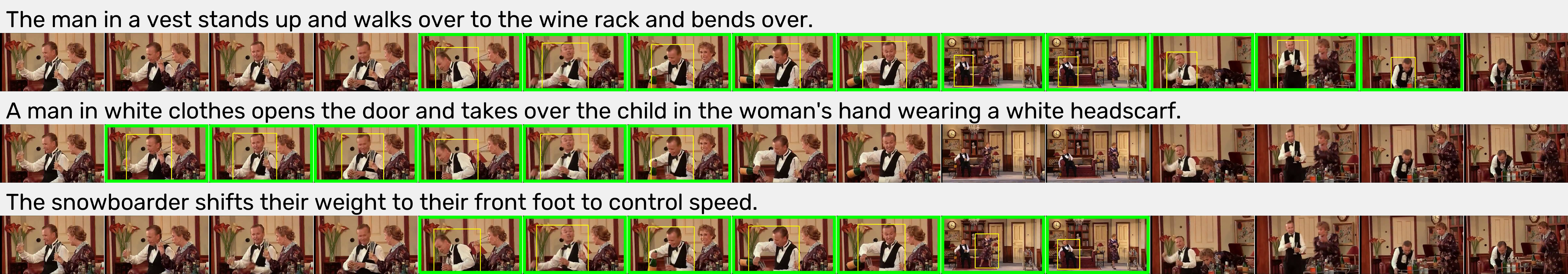}
        \caption{Video ID: 102\_VsYPP2I0aUQ}
    \end{subfigure}
    \begin{subfigure}[b]{1.0\textwidth}
        \centering
        \includegraphics[width=1\textwidth]{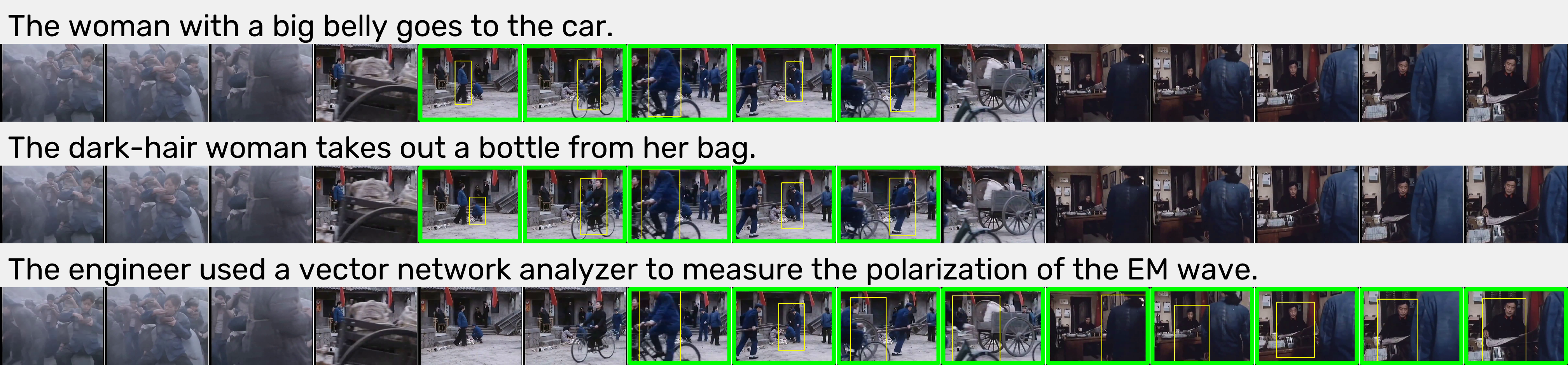}
        \caption{Video ID: 111\_Uw7387tc9PU}
    \end{subfigure}
    \captionsetup{width=\textwidth}
    \caption{Qualitative results of spatio-temporal grounding prediction of TA-STVG model (weights shared by its authors) on the example HCSTVG-v2 videos. The green border marks temporal prediction frame, the yellow box stands for spatial prediction. Input queries are in order: (1) positive (2) negative in-domain (3) negative out-of-domain.}
    \label{fig:qual1}
\end{figure*}

To better illustrate the behavior under irrelevant textual input, we present qualitative examples of TA-STVG model predictions on a five example videos from HCSTVG-v2 dataset. We evaluated them on the original positive query, the negative in-domain query and the negative out-of-domain query. The examples presented in Fig.~\ref{fig:qual1} illustrate that, even when the query is semantically unrelated to the target video, the model may still predict visually salient entities or dataset-prior-consistent regions instead of rejecting the query.

\newpage\newblock

\end{document}